\documentclass{article}
\usepackage{hyperref}
\usepackage{graphicx}
\usepackage{spconf,epsfig, amsfonts,amssymb}
\usepackage[nosumlimits]{amsmath}

\let\OLDthebibliography\thebibliography
\renewcommand\thebibliography[1]{
  \OLDthebibliography{#1}
  \setlength{\parskip}{0pt}
  \setlength{\itemsep}{0pt plus 0.3ex}
}

\begin{document}\sloppy

\def\x{{\mathbf x}}
\def\L{{\cal L}}

\newcommand{\figref}[1]{{Fig.~\ref{#1}}}
\newcommand{\tabref}[1]{{Tab.~\ref{#1}}}
\newcommand{\argmax}{\mathop{\rm arg~max}\limits}
\newcommand{\argmin}{\mathop{\rm arg~min}\limits}
\newcommand{\indicator}{{\mbox{1}\hspace{-0.25em}\mbox{l}}}
\newcommand{\pdfrac}[2]{{\frac{\partial {#1}}{\partial {#2}}}}
\newcommand{\inputfig}[5][]{
\begin{figure#1}[tb]
\centering
\includegraphics[width=#2\linewidth]{#3}
\caption{#4}
\label{#5}
\end{figure#1}
}

\title{Learning Robust Convolutional Neural Networks \\ with Relevant Feature Focusing via Explanations}
%
\name{Kazuki Adachi and Shin'ya Yamaguchi}
\address{Computer and Data Science Laboratories, NTT Corporation, Japan \\
\{kazuki.adachi, shinya.yamaguchi\}@ntt.com}

\maketitle

\begin{abstract}
Existing image recognition techniques based on convolutional neural networks (CNNs) basically
 assume that the training and test datasets are sampled from i.i.d distributions.
However, this assumption is easily broken in the real world because of the distribution shift that occurs when the co-occurrence relations between objects and backgrounds in input images change.
Under this type of distribution shift, CNNs learn to focus on features that are not task-relevant, such as backgrounds from the training data, and degrade their accuracy on the test data.
To tackle this problem, we propose {\it relevant feature focusing (ReFF)}.
ReFF detects task-relevant features and regularizes CNNs via explanation outputs (e.g., Grad-CAM).
Since ReFF is composed of post-hoc explanation modules, it can be easily applied to off-the-shelf CNNs.
Furthermore, ReFF requires no additional inference cost at test time because it is only used for regularization while training.
We demonstrate that CNNs trained with ReFF focus on features relevant to the target task and that ReFF improves the test-time accuracy.

\end{abstract}

\section{Introduction}
\label{sec:intro}
Convolutional neural networks (CNNs) have achieved high accuracy in image recognition tasks~\cite{lecun_cnn,alexnet} under the assumption that the  training and test datasets are sampled from i.i.d distributions.
However, in the real world, this assumption may not be valid and the distributions easily shift
because of changing contexts in images such as in the co-occurrences between backgrounds and objects~\cite{Singh_CVPR_2020,action_recognition_bias_neurips2019,bahng2020learning}.
In such cases, the accuracy of the CNNs is degraded.

One of the causes of distribution shift is using a biased procedure to collect the training dataset.
Practitioners often collect training data from biased contexts (e.g., backgrounds) due to budget limitations.
A training dataset made within such limitations may have {\it spurious features}, that are correlated with labels only in the training dataset even though they are originally unrelated to the task~\cite{NEURIPS2020_f1298750,pmlr-v139-zhou21g}.
Spurious features can easily vanish or mutate in a test environment where the context changes.
If CNNs learn spurious features, their accuracy may be degraded at test time.
In such a case, the reliability of the CNNs also deteriorates since they focus on features unrelated to the task.
This problem frequently occurs in image recognition because the context may change after the models are deployed~\cite{Singh_CVPR_2020,bahng2020learning}, as illustrated in \figref{fig:sheep_example}.
Thus, to train robust and reliable CNNs, it is important to focus on {\it task-relevant features} that are originally related to the task and are invariant to the training and test distributions~\cite{Singh_CVPR_2020,Rieger2020}.

\inputfig{0.73}{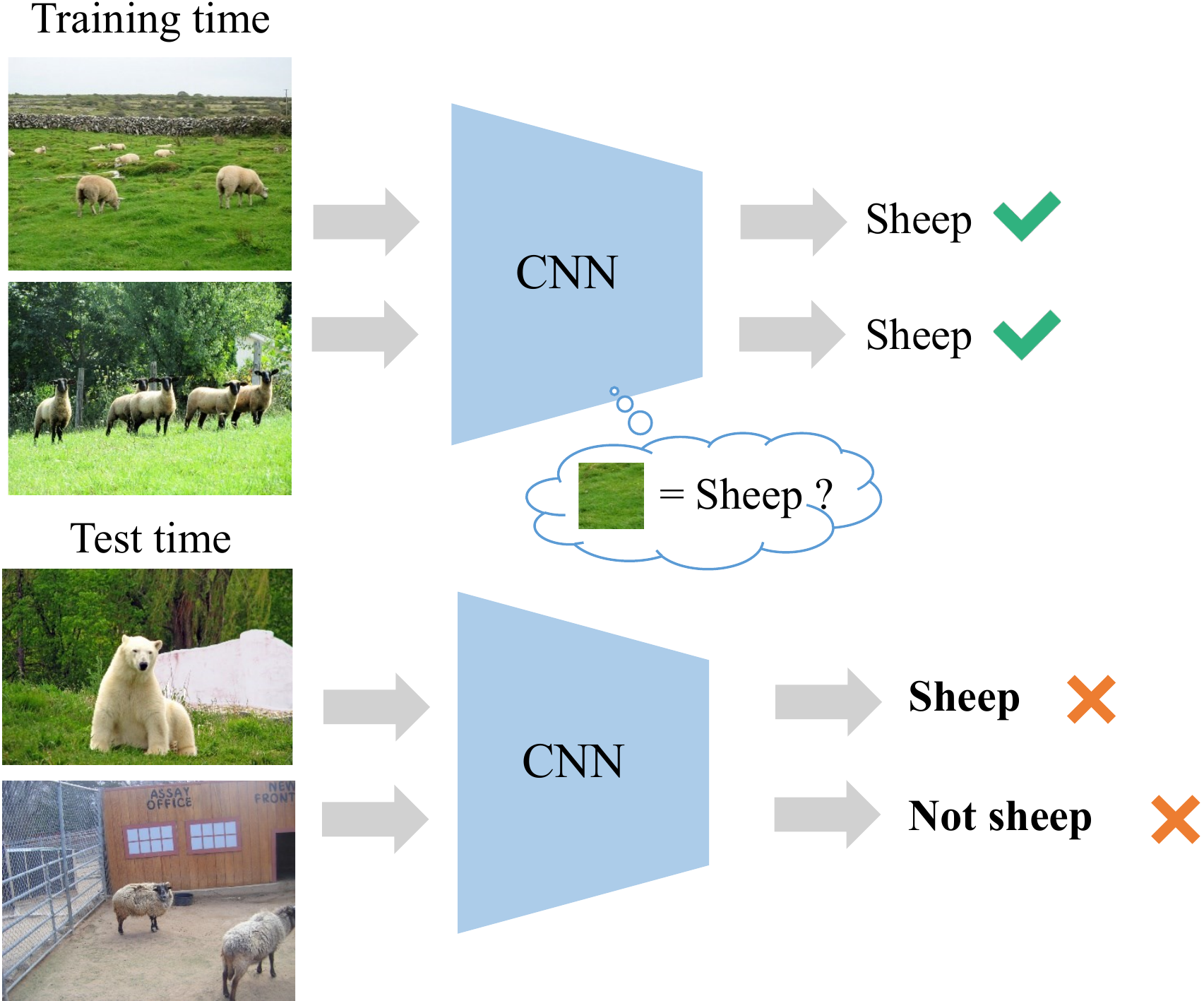}{
Example of spurious features.
CNNs mistakenly learn the features of grassy fields to be `sheep' because the training images labeled `sheep' tend to contain grassy fields.
But at test time, images of non-sheep in grassy fields are classified as `sheep' and vice versa.
The images are sampled from the AwA dataset~\cite{animals-with-attributes}.
}{fig:sheep_example}

To avoid degradation of accuracy caused by spurious features and improve the reliability at test time, we propose {\it relevant feature focusing (ReFF)}.
ReFF consists of an {\it explanation regularizer} and a {\it pseudo-annotator}.
The explanation regularizer constrains the CNN classifiers to focus on task-relevant features and avoid spurious features 
by minimizing the gap between the explanations of the classifiers and regional annotations, which indicate the regions of task-relevant features in the training images.
The explanation regularizer can be applied to off-the-shelf CNN classifiers because it uses a post-hoc explanation module for computing the regularization term.
Although it can make CNNs focus on task-relevant features, it requires the regional annotations to be manually created for each training image.
Here, we use the pseudo-annotator to reduce the number of regional annotations.
The pseudo-annotator detects regions of task-relevant features from the input images and generates regional annotations 
with an encoder-decoder network inspired by image translation~\cite{pix2pix}.
We experimentally confirmed that a pseudo-annotator trained on only several manual regional annotations can improve the accuracy of the classifiers.
We conducted experiments on artificial (Textured MNIST) and real world datasets (ISIC2017~\cite{codella2018skin}, Oxford-IIIT Pets~\cite{oxford_pets}) 
and confirmed that spurious features affect CNNs especially when the datasets have diverse task-relevant features or the task-relevant features are small in addition to there being a strong correlation between spurious features and labels.
Moreover, we found that CNN classifiers trained with ReFF had higher test accuracy in the presence of spurious features by focusing on task-relevant features.

\section{Related Work}\label{sec:related_works}
{\bf Domain adaptation:} Domain adaptation (DA)~\cite{csurka2017domain} is a popular technique for dealing with the distribution shift at test time.
It helps models to learn features that are invariant to the source and target datasets.
However, DA requires a target dataset collected from the test environment for training.
In practice, it is difficult to access the test environment before deploying models.
Contrastively, our goal is to train models that generalize to the test environment without accessing it. \\
{\bf Regularizing via attention maps:} In the context of explainable AI, several methods to visualize the predictions of CNNs have been developed~\cite{grad_cam,cam,Fukui_2019_CVPR,sundararajan2017axiomatic}.
These methods have also been used for regularizing CNNs to focus on task-relevant features and to be robust against spurious features~\cite{embedding_human_knowledge,Singh_CVPR_2020,Rieger2020}.
In contrast to DA methods, these regularization methods help CNNs generalize to unseen test datasets without accessing test environments at training time by focusing on task-relevant features.
Mitsuhara et al.~\cite{embedding_human_knowledge} used a specialized CNN architecture called ABN~\cite{Fukui_2019_CVPR} for regularizing CNNs via  attention maps.
They minimized the mean squared errors between  attention maps generated by ABN and manual regional annotations.
However, since ABN is composed of specialized architectures, it is hard to apply to off-the-shelf architectures.
On the other hand, several studies have used post-hoc explanation modules for generating attention maps~\cite{Singh_CVPR_2020,Rieger2020}.
Although the post-hoc modules can be adapted to off-the-shelf architectures, these methods require multi-class labels or numerous additional regional annotations (e.g., 30,917 regional annotations in~\cite{embedding_human_knowledge}).
ReFF uses a post-hoc explanation module and can be applied to single-class labels.
Here, with an eye to keeping flexibility, we reduced the number of additional annotations by using a pseudo-annotator built with an image translation model, which is trainable on a smaller number of annotations in Sec.~\ref{ssec:pseudo_annotation}.

\section{Focusing on Task-relevant Feature} \label{sec:proposed_method}
\inputfig{0.9}{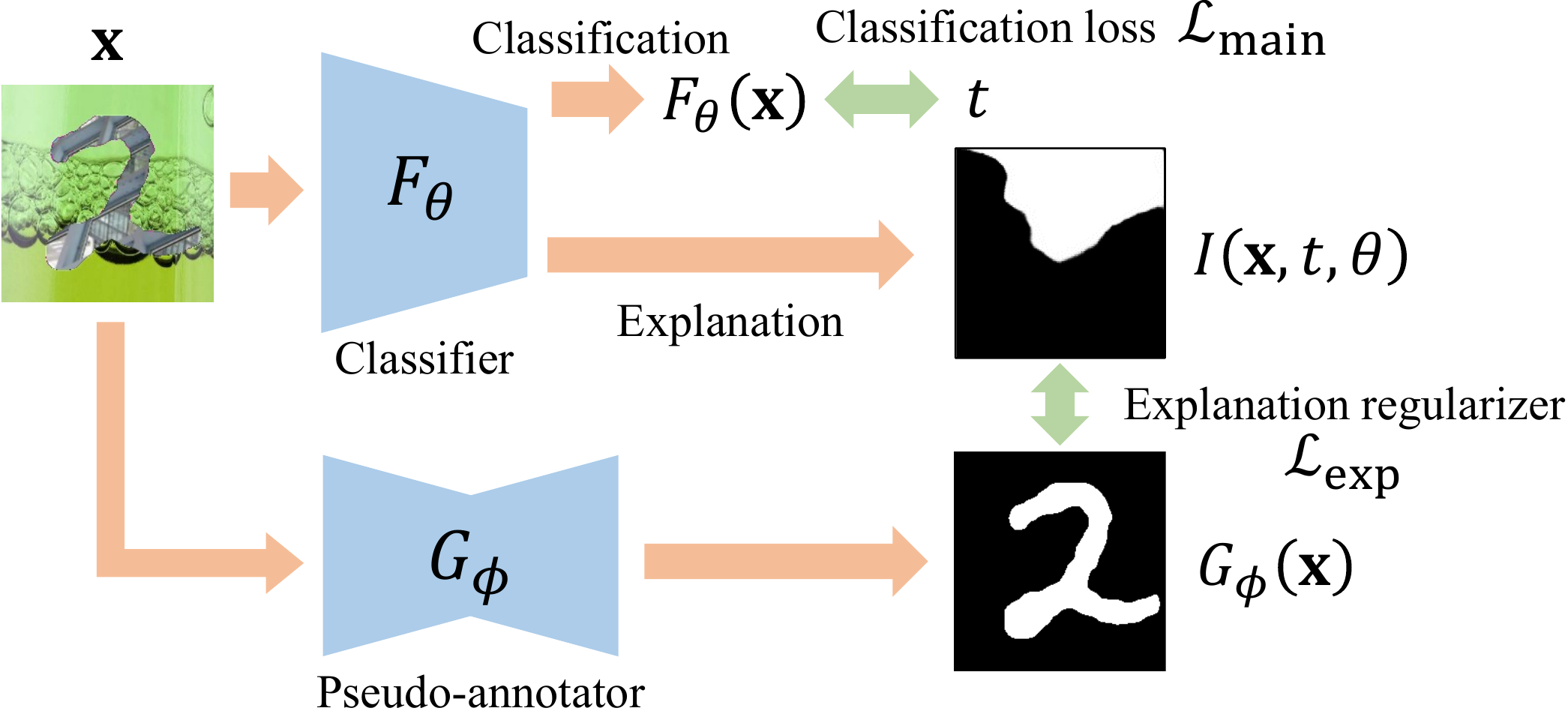}{
Overview of relevant feature focusing (ReFF).
}{fig:iff_overview}
In this section, we introduce ReFF, which makes CNNs focus on task-relevant features instead of spurious ones.
As shown in \figref{fig:iff_overview}, ReFF is composed of an {\it explanation regularizer} and a {\it pseudo-annotator}.
The explanation regularizer minimizes the gap between the explanations of the CNNs and the regional annotations by regularizing multiple intermediate layers
in such a way that the CNNs focus on task-relevant features.
The pseudo-annotator is an auxiliary model used at training time to detect task-relevant features  
and generate regional annotations instead of manually collecting them.
In practice, a pseudo-annotator can be trained on a reasonable number (hundreds or even dozens) of regional annotations.

\subsection{Explanation Regularizer}\label{ssec:explanation_regularizer}
ReFF regularizes the CNN to focus on task-relevant features via the explanations of the CNN~\cite{grad_cam,cam,sundararajan2017axiomatic}.
Explanations were originally designed to interpret the predictions of CNNs.
Most explanation methods generate heat maps that indicate regions in the input images that the CNNs focus on.
We use these heat maps for regularizing the CNNs.
In paticular, ReFF uses Grad-CAM~\cite{grad_cam} as an explanation because it is easy to implement 
and can be computed from arbitrary intermediate layers of CNNs.
We will denote Grad-CAM computed from layer $l$ by $I_l(\mathbf{x},y,\theta) \in \mathbb{R}_+^{H^l\times W^l}$, where $\mathbf{x} \in \mathbb{R}^{3\times H\times W}$ is an input image, $y$ is the class label for $\mathbf{x}$, and $\theta$ consists of parameters of the CNN classifier $F_\theta$.
$I_l(\mathbf{x},y,\theta)$ indicates the important region of $\mathbf{x}$ to be classified to $y$ by $F_\theta$.
Grad-CAMs are produced by calculating the sum of the feature maps $A^l\in \mathbb{R}^{K^l\times H^l\times W^l}$ weighted by their gradients:
\begin{eqnarray}
I_l(\mathbf{x},y,\theta) &=& \text{ReLU}\left( \sum_{k=1}^{K^l} \alpha_k^l A^l_k \right), \\
\alpha_k^l &=& \frac{1}{H^l W^l} \sum_{i,j} \frac{\partial F_\theta (\mathbf{x})_y}{\partial A^l_{k,i,j}}.
\end{eqnarray}
The explanation regularizer matches $I_l(\mathbf{x},y,\theta)$ and the regional annotations of $\mathbf{x}$, denoted by $\mathbf{s}$, for each layer as follows:
\begin{equation}
\mathcal{L}_\text{ReFF} = \mathbb{E}_{\mathbf{x},y,\mathbf{s}}\left[ \sum_{l=1}^L w_l \left\| I_l'(\mathbf{x},y,\theta)\odot (\mathbf{1}-\mathbf{s}) \right\|_2^2 \right],
\label{eq:iff_loss}
\end{equation}
where $\odot$ is the element-wise product, $w_l$ is a hyperparameter to determine the weight for layer $l$, 
and $\mathbf{s} \in [0,1]^{H\times W}$ is a regional annotation that has the same resolution as $\mathbf{x}$.
$\mathbf{s}$ is 1 for regions of task-relevant features and 0 for other regions.
$I_l'(\mathbf{x},y,\theta)_{i,j}=I_l(\mathbf{x},y,\theta)_{i,j} / \| I_l(\mathbf{x},y,\theta) \|_1$ is the normalized explanation.
Since each $I_l(\mathbf{x},y,\theta)$ has a different resolution, we resize it to the same resolution as $\mathbf{s}$ by linear interpolation.
While previous studies~\cite{Rieger2020,embedding_human_knowledge} strictly match explanations and regional annotations simply by minimizing their L1 or L2 error, 
we minimize the norm of the explanations only outside the regions of the task-relevant features to enable the CNNs to implicitly learn the importance of the task-relevant features inside the regions.
In fact, we found that aligning $I(\mathbf{x},y,\theta)$ and $\mathbf{s}$ by using the L1 or L2 error instead of $\mathcal{L}_\text{ReFF}$ has a negative effect (see Sec. B of the supplementary materials) because each pixel inside the regions of the task-relevant features has a different importance, while $\mathbf{s}$ usually contains binary values that does not represent the importance.
In addition, we designed the explanation regularizer to penalize multiple layers of CNNs 
because each layer extracts the features of different levels, from low-level textures to high-level discriminative objects~\cite{deconvnet,NIPS2014_375c7134}.
By regularizing the explanations of multiple layers simultaneously, $F_\theta$ can learn various levels of task-relevant feature.
We add $\mathcal{L}_\text{ReFF}$ to main objective $\mathcal{L}_\text{main}$, the whole objective is $\mathcal{L}=\mathcal{L}_\text{main}+\lambda \mathcal{L}_\text{ReFF}$, 
where $\lambda$ is a hyperparameter to balance the main loss term and the regularization term.

\subsection{Pseudo-annotator}
The explanation regularizer in Eq.~\eqref{eq:iff_loss} requires a regional annotation $\mathbf{s}$ for each image.
To reduce the number of regional annotations, we incorporate a pseudo-annotator $G_\phi: \mathbb{R}^{3\times H\times W} \mapsto [0,1]^{H\times W}$, which generates pseudo-annotations of $\mathbf{s}$ for the explanation regularizer.
We regard this pseudo-annotation generation task as an image-translation that maps an input image to a pseudo-annotation.
We use pix2pix~\cite{pix2pix} for the image-translation.
Pix2pix is a supervised image-translation method.
The experimental results reported in~\cite{pix2pix} show that pix2pix can be trained on a small dataset consisting of hundreds of images.
We expect that the pseudo-annotator can be trained with the same order of images.
After training $G_\phi$, a pseudo-annotation $G_\phi (\mathbf{x})$ is used for samples that do not have $\mathbf{s}$ as Eq.~\eqref{eq:reff_loss_with_pseudo-annotator} instead of Eq.~\eqref{eq:iff_loss}.
\begin{equation}
\mathcal{L}_\text{ReFF} \!=\! \mathbb{E}_{\mathbf{x},y} \! \left[ \sum_{l=1}^L \! w_l \left\| I_l'(\mathbf{x},y,\theta)\!\odot\! (\mathbf{1}\!-\!G_\phi(\mathbf{x})) \right\|_2^2 \right].
\label{eq:reff_loss_with_pseudo-annotator}
\end{equation}

\section{Experiments}\label{sec:experiment}
We identify when spurious features affect CNNs and confirm that ReFF helps CNNs to learn task-relevant features and avoid the effect of spurious features.
First, we examine the effect of the explanation regularizer on Textured MNIST directly by using regional annotations 
without the pseudo-annotator.
Then, we test the pseudo-annotator on all datasets to evaluate the total effect of ReFF.

\subsection{Datasets}
We used three datasets: Textured MNIST, ISIC2017~\cite{codella2018skin}, and Oxford-IIIT Pet (Pets)~\cite{oxford_pets}.
Textured MNIST is an artificial dataset we made, the others are real-world datasets.
\\
{\bf Textured MNIST} is composed of MNIST~\cite{mnist} and DTD~\cite{cimpoi14describing} for the purpose of assessing the effect of spurious features by controlling co-occurrences of digits and background textures.
We placed a digit image from MNIST in a $224\times 224$ canvas and put different texture images from DTD on each of the digit and the background.
The task is to classify the images into 10 digit classes without being affected by background textures.
We simulated spurious features by fixing the classes of background textures corresponding to each digit class.
Thus, backgrounds are spurious features correlated with the digit classes.
We controlled the degree of effect of the spurious features with a background texture randomness $p$ and a digit texture randomness $q$.
We selected a background texture from a random class with probability $p$ 
and selected another background texture from the fixed class corresponding to the digit class with probability $1-p$.
We also selected the textures of the digits in the same way with probability $q$.
When $p$ is small, the images in the dataset have strong spurious features that CNNs will likely learn even though they are originally not related to digit classification.
A larger $q$ generates a wider diversity of task-relevant features and makes the spurious features more discriminative than the task-relevant ones.
In the test dataset, $p$ is set to 1 so that the correlation between the spurious features and the labels is broken.
Further, we controlled the size of the digits to evaluate the effect of the size of the task-relevant features.
\\
{\bf ISIC2017}~\cite{codella2018skin} is a dataset consisting of three classes of skin lesion image.
Some of the images contain not only lesions but also artifacts such as colorful patches, scales, and  markers~\cite{mahbod2019skin}.
These artifacts can be regarded as spurious features.
\\
{\bf Oxford-IIIT Pet (Pets)}~\cite{oxford_pets} is a dataset consisting of images of cats and dogs, classified into 37 classes.
We selected this dataset as an example of a natural dataset that does not have trivial spurious features, unlike ISIC2017.

Sec. A of the supplementary materials provides details on and examples from the datasets.

\subsection{Training Details}
{\bf Classification: }We used ResNet-50~\cite{resnet} pretrained on ImageNet~\cite{lsvrc2012} for the classifier.
We applied the explanation regularizer to the outputs of the blocks of {\tt layer1} to {\tt layer4}.
We set the weights for each layer in Eq.~\eqref{eq:iff_loss} $\lambda=1$ and $(w_1,w_2,w_3,w_4) = (15,60,250,1000)$, 
where $\lambda$ was set as such to make $\mathcal{L}_\text{ReFF}$ and $\mathcal{L}_\text{main}$ roughly of the same order
and $w_i$ is for balancing the importance of the layers.
For $w_i$, we found that assigning larger weights to higher layers produced better results by performing a grid search.
For the objective, we implemented $\mathcal{L}_\text{main}$ with cross entropy. \\
{\bf Pseudo-annotator: }We used U-Net~\cite{u-net} for the pseudo-annotator $G_\phi$ in the same way as~\cite{pix2pix}.

The detailed settings are in Sec. D of the supplementary materials.

\begin{table}[tb]
\centering
\caption{Test accuracy [{\%}] on Textured MNIST.
For each condition, we ran the training three times with different random seeds, the standard deviations are shown together.
}
\label{tab:textured_mnist_accuracy}
\includegraphics[width=1\linewidth]{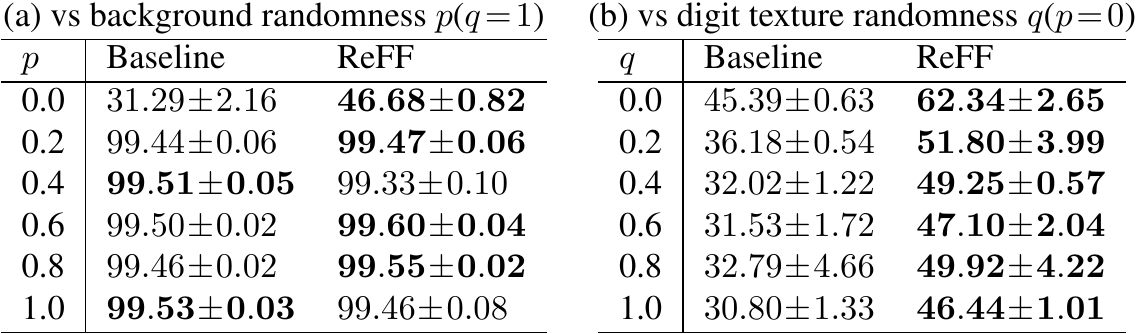}
\end{table}

\begin{figure}[tb]
\centering
\begin{tabular}{ccc}
\includegraphics[height=6.7em]{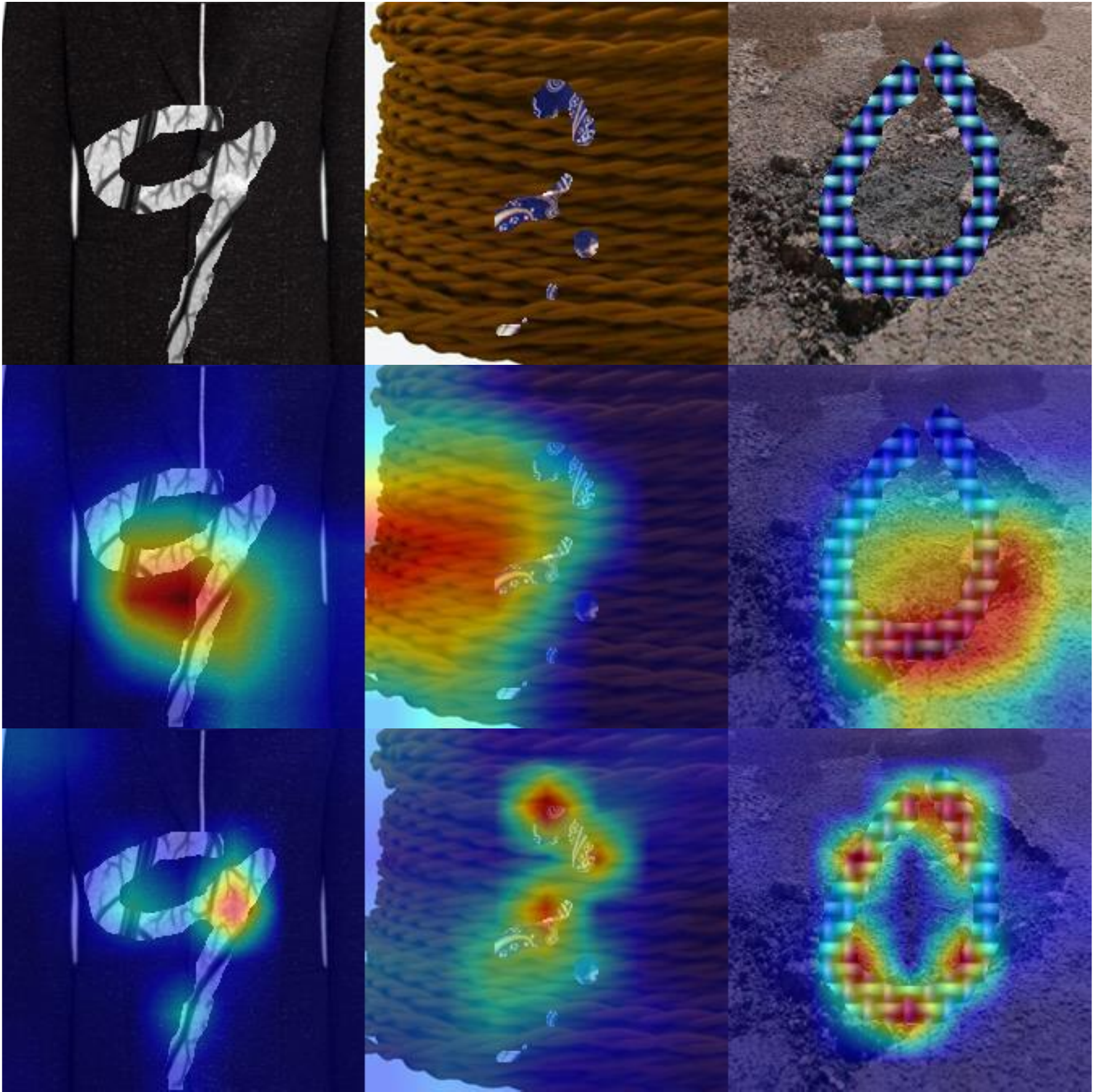}
&
\includegraphics[height=6.7em]{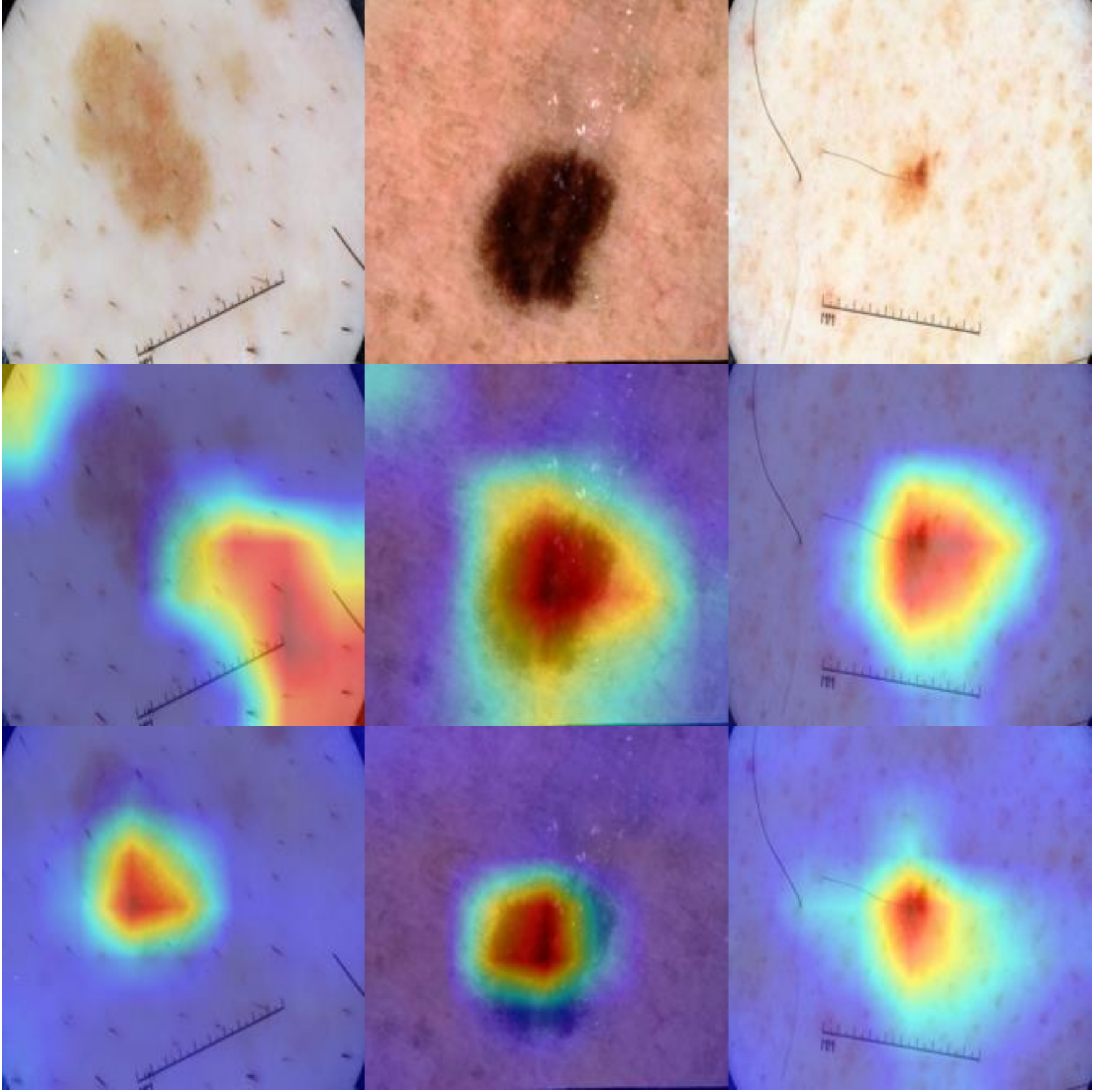}
&
\includegraphics[height=6.7em]{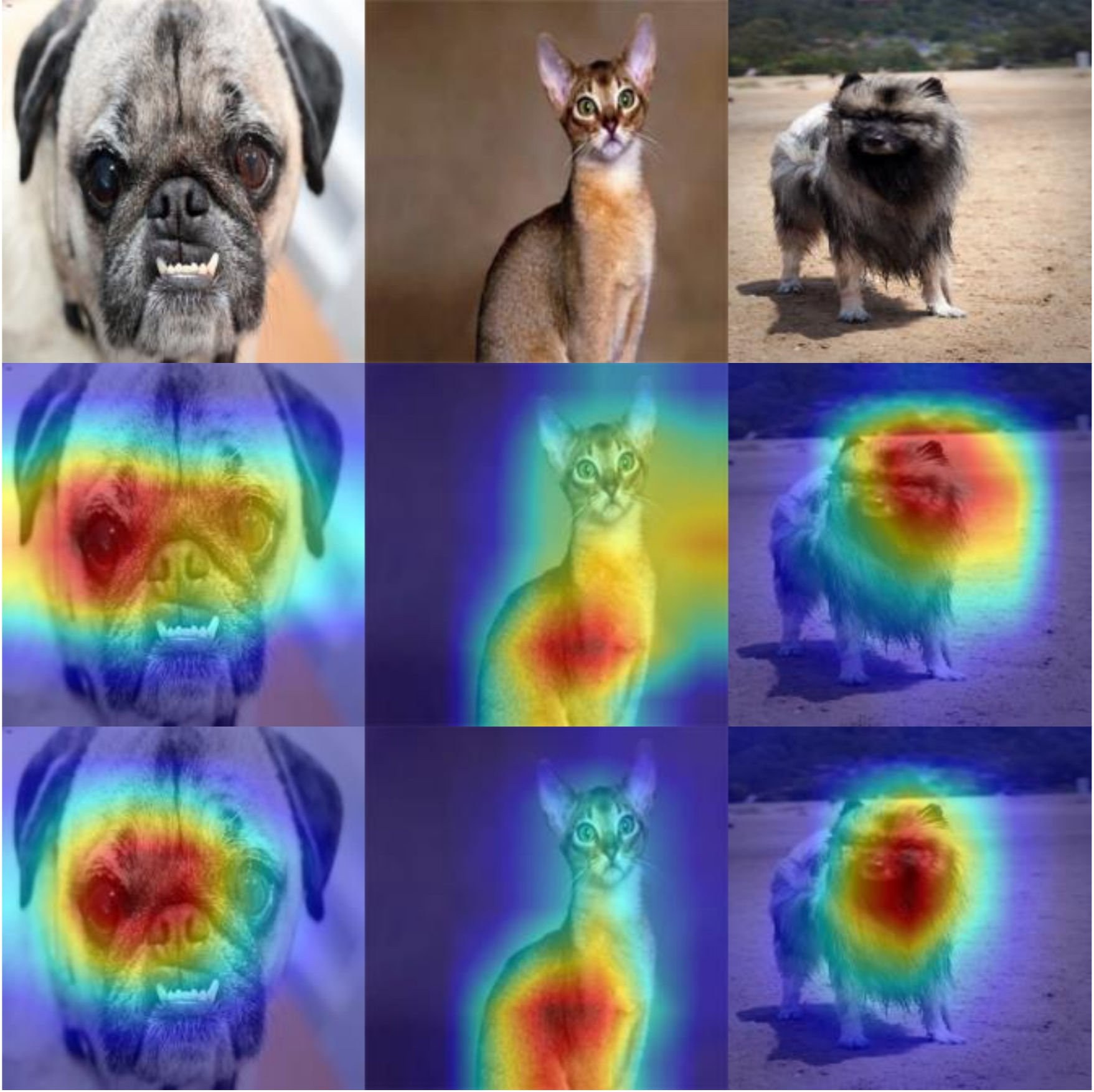} \\
{\ninept (a) Textured MNIST} &
{\ninept (b) ISIC2017} &
{\ninept (c) Pets}
\end{tabular}

\caption{Grad-CAMs computed from the outputs of {\tt layer4}.
The red regions indicate where the classifiers attach importance in order to make a prediction, the blue regions are the opposite.
The upper, middle, and lower rows show the inputs, baseline, and ReFF, respectively.}
\label{fig:cam_examples}
\end{figure}

\subsection{Effect of Spurious Features} \label{ssec:exp_spurious_features}
We evaluate the accuracy of the CNNs under various degrees of spurious features.
We trained the classifiers on Textured MNIST by varying the background randomness $p$ and the digit texture randomness $q$ in the training dataset.
We fixed the digit size to 223.
We compared ReFF with a baseline that minimizes only the cross entropy loss.
First, we fixed $q=1$ and varied $p$ to evaluate the effect of spurious features.
\tabref{tab:textured_mnist_accuracy} (a) shows the test accuracies with respect to $p$.
When $p>0$ in the training dataset, both the baseline and ReFF achieve high accuracy.
However, when $p=0$ (fixed background textures), the accuracy drops because the correlation between the spurious features and labels is strong in the training dataset.
Here, ReFF alleviated the accuracy drop by preventing the classifiers from learning the spurious features.
Next, we set $p=0$ and varied the digit texture randomness $q$ to evaluate the effect of the variety of task-relevant features under strong spurious features.
In this setting, when $q$ is higher, the test is more difficult because the spurious features (fixed textures of backgrounds) are more discriminative than the task-relevant features (randomly sampled textures of digits).
\tabref{tab:textured_mnist_accuracy} (b) shows the results.
Here, the baseline model learned spurious features and the test accuracy deteriorated as $q$ increased.
On the other hand, ReFF achieved a higher test accuracy than the baseline model by about 15{\%}pt in all cases.
This suggests that ReFF can learn task-relevant features even when strong spurious features exist in the training dataset.
The effect of ReFF can also be seen in the explanations of the models. 
\figref{fig:cam_examples} (a) shows the Grad-CAMs of the test data generated from the classifier trained on $p=0$.
As shown in the middle row, the baseline model focused on background textures that are unrelated to digit classification.
This is because the background textures are discriminative features only in the training dataset.
In contrast, as shown in the bottom row, the CNN trained with ReFF properly focused on digits.
We also evaluate ReFF with additional cases (Sec. C of the supplementary materials).

\subsection{Effect of the Size of the Task-Relevant Features}\label{ssec:exp_size}
\begin{table}[tb]
\centering
\caption{Test accuracy [\%] vs. digit size on Textured MNIST.}
\label{tab:size-hard_accuracy}
\includegraphics[width=0.8\linewidth]{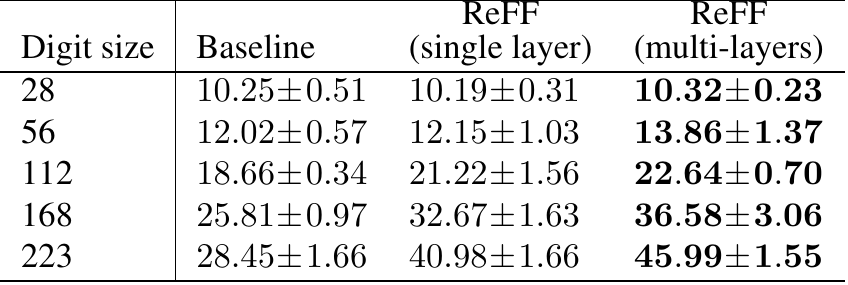}

\end{table}

We varied the size of the digits of Textured MNIST to show that applying ReFF regularization to multiple layers simultaneously 
enables various sizes of task-relevant feature to be learned.
We set $(p,q)=(0,1)$ for the training dataset and $(p,q)=(1,1)$ for the test dataset, 
where the effect of spurious features is the strongest.
We placed the digits at random positions in each image.
\tabref{tab:size-hard_accuracy} shows the test accuracies.
We compared ReFF with its variant, which uses only {\tt layer4} in Eq.~\eqref{eq:iff_loss} with $\lambda=1000$ and $w_\text{layer4}=1$.
When the digit size $\geq 112$, ReFF with multi-layers and ReFF with a single layer improved the test accuracy.
But ReFF with multi-layers was the more accurate.
Especially for a digit size of 56, it showed an accuracy improvement over the baseline 
while the accuracy of ReFF with a single layer stayed almost the same as that of the baseline. 
This means that regularizing lower layers can capture finer task-relevant features better than regularizing only a higher layer.
Since the resolution of the feature maps of higher layers become coarser by the stride of the convolutions, capturing fine task-relevant features with only a single high layer is difficult. 
On the other hand, for a digit size of 28, the accuracies of the baseline and ReFF are both deteriorated and their difference was smaller.
This means that ReFF works effectively when the regions of task-relevant features are large.
When the task-relevant features are small and buried in spurious features (backgrounds) like in this case, the accuracy of ReFF gets closer to that of the baseline.
Note that this is an artificial extreme case that hardly occurs in natural datasets, as evidenced by the experiments on ISIC2017 and Pets described below.

\subsection{Effect of Adding the Pseudo-annotator}\label{ssec:pseudo_annotation}
\begin{table}[tb]
\centering
\caption{Test accuracy [\%] vs. training subset size $n$ in the case of using the pseudo-annotator.
}
\label{tab:pseudo-annotator_accuracy}
\begin{tabular}{c}
{\ninept \shortstack{
(a) Comparison of baseline and ReFF on three datasets. \\
The best accuracy of all and the best accuracy of ReFF are \\
respectively in {\bf bold} and \underline{underlined}.
}}
\\
\includegraphics[width=0.7\linewidth]{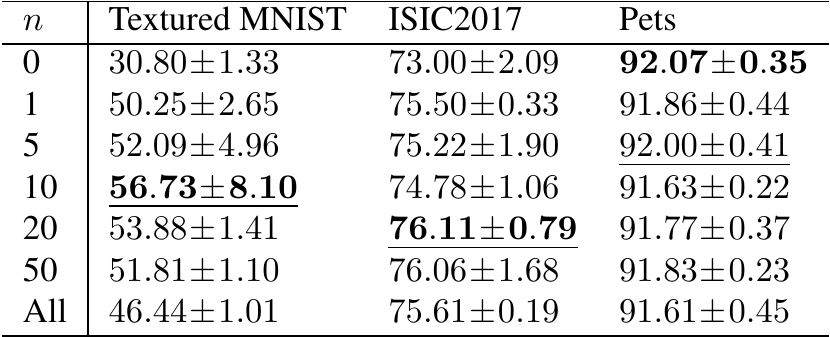}
\\
{\ninept \shortstack{
(b) Comparison of ReFF and ABN-based regularization~\cite{embedding_human_knowledge}.
}}
\\
\includegraphics[width=0.5\linewidth]{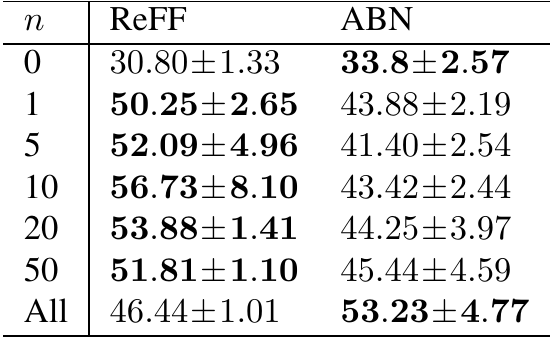}
\end{tabular}
\end{table}

We evaluate the effect of incorporating the pseudo-annotator $G_\phi$ in terms of the number of regional annotations.
First, we trained $G_\phi$ on subsets of the training dataset with regional annotations $\mathbf{s}$.
We varied the size of the subsets of the training dataset, which means the number of manually collected regional annotations. 
We made the subsets by sampling $n\in \{ 0,1,5,10,20,50, \text{All} \}$ images {\it per class} from the training dataset.
Then, we trained the CNN classifiers with the trained $G_\phi$.
In this step, we used all of the training data, but gave regional annotations $\mathbf{s}$ for only the subset selected in the former step.
For the samples with $\mathbf{s}$, we calculated the explanation regularizer term as Eq.~\eqref{eq:iff_loss}.
For the rest, we used  $G_\phi(\mathbf{x})$ instead of $\mathbf{s}$ as Eq.~\eqref{eq:reff_loss_with_pseudo-annotator}.
The detailed training procedure is described in Alg.~1 in the supplementary materials.
Note that $n=0$ means that no regional annotations are used (i.e., baseline), and $n=\text{`All'}$ means that ReFF uses regional annotations directly for all training data without the pseudo-annotator.
For Textured MNIST, we set $(p,q)=(0,1)$ for the training dataset, where the effect of spurious features is the strongest.
For the test dataset, we set $(p,q)=(1,1)$, where spurious features do not exist.
\tabref{tab:pseudo-annotator_accuracy} (a) shows the test accuracy on the three datasets trained with and without the pseudo-annotator.
For Textured MNIST, ReFF significantly improved the test accuracy in comparison with the baseline ($n=0$) in spite of the strong spurious features in the training dataset.
Even when $n=1$, where the pseudo-annotator was trained on only $10~(=1~\text{sample per class}\times 10~\text{class})$ samples, ReFF exceeded the baseline by about 20{\%}pt.
More interestingly, the $1 \leq n \leq 50$ cases outperformed the $n=\text{`All'}$ case, which uses regional ones directly.
For this reason, we consider that pseudo-annotations are more informative than regional annotations.
While regional annotations have binary values indicating the regions of task-relevant features, 
the pseudo-annotator outputs `soft' pseudo-annotations that represent the importance of each pixel.
For ISIC2017, ReFF also improved accuracy compared with the baseline even when $n=1$, 
where the pseudo-annotator was trained on only 3 samples.
We can also find cases in which it outperforms the $n=\text{`All'}$ case.
As shown by the Grad-CAMs in \figref{fig:cam_examples} (b), the ReFF model focused on lesions correctly while the baseline model highlighted areas outside the lesions.
For Pets, ReFF achieved competitive accuracy with the baseline especially when $n=5$.
This is because we fine-tuned the ImageNet-pretrained model.
Since ImageNet contains images of dogs and cats, 
the ImageNet-pretrained model has already learned features of dogs and cats.
Thus, the baseline model learnd task-relevant features as well as ReFF with this dataset.
On the other hand, ReFF improved explainability, as shown in \figref{fig:cam_examples} (c).
The baseline model sometimes focused on backgrounds, which were originally independent of the classification,
whereas the ReFF model highlighted animals more accurately.
Thus, the ReFF model generated clear and explanations, which improves reliability.

We also compared ReFF with the ABN-based regularization method~\cite{embedding_human_knowledge} (denoted by ABN) on Textured MNIST.
\tabref{tab:pseudo-annotator_accuracy}~(b) shows the result.
ABN uses only manual regional annotations, so we also combined it with our pseudo-annotator.
When $n=\text{`All'}$, where regional annotations are given for all training data, 
ABN exceeded ReFF because ABN learned accurate attentions with its specialized module called attention branch that is independent of the predictions.
On the other hand, when $n\leq 50$ where pseudo-annotations are sometimes inaccurate, ReFF outperformed ABN.
This is because our explanation regularizer is robust to corruption of the pseudo-annotations because it 
suppresses explanations only outside the regions of task-relevant features, as mentioned in Sec.~\ref{ssec:explanation_regularizer}.
Contrastively, since ABN minimizes the L2 error between regional annotations and attentions, the model attempts to align its attentions with the corrupted pseudo-annotations strictly and fails to focus on task-relevant features.
We also compared variants of ReFF and visualized the pseudo-annotations, Grad-CAMs, and attentions of ABN (see Sec. B of the supplementary materials).

\section{Conclusion}
We developed ReFF for training CNN classifiers that are robust to changes in spurious features.
ReFF regulates the explanations of the classifiers to learn task-relevant features.
In addition, it detects task-relevant features and generates pseudo-annotations with a pseudo-annotator trained on a small number of regional annotations.
We experimentally showed that ReFF efficiently learns task-relevant features from training datasets even when strong spurious features exist.
ReFF made accurate predictions on the test datasets and improved explanations even when the spurious features vanished or mutated.
Moreover, we found that pseudo-annotations further improved the test accuracy beyond that of using regional annotations.
In the future, we will explore the effect of the pseudo-annotator and improve ReFF to allow models to learn task-relevant features more efficiently.

\bibliographystyle{IEEEbib}
{\ninept \bibliography{main}}

\end{document}


\sloppy

\def\x{{\mathbf x}}
\def\L{{\cal L}}

\newcommand{\figref}[1]{{Fig.~\ref{#1}}}
\newcommand{\tabref}[1]{{Tab.~\ref{#1}}}
\newcommand{\argmax}{\mathop{\rm arg~max}\limits}
\newcommand{\argmin}{\mathop{\rm arg~min}\limits}
\newcommand{\indicator}{{\mbox{1}\hspace{-0.25em}\mbox{l}}}
\newcommand{\pdfrac}[2]{{\frac{\partial {#1}}{\partial {#2}}}}
\newcommand{\inputfig}[5][]{
\begin{figure#1}[tb]
\centering
\includegraphics[width=#2\linewidth]{#3}
\caption{#4}
\label{#5}
\end{figure#1}
}

\title{Supplementary Materials for \\
Learning Robust Convolutional Neural Networks \\ 
with Relevant Feature Focusing via Explanations}
%
\name{Kazuki Adachi and Shin'ya Yamaguchi}
\address{Computer and Data Science Laboratories, NTT Corporation, Japan \\
\{kazuki.adachi, shinya.yamaguchi\}@ntt.com}

\maketitle

This manuscript is the supplementary materials of ``Learning Robust Convolutional Neural Networks with Relevant Feature Focusing via Explanations''.
Sec.~\ref{sec:datasets} provides the details on the datasets used in the experiments of the main paper.
Sec.~\ref{sec:explanation_regularizer} provides the additional experiments with variants of ReFF.
Sec.~\ref{sec:experiment_relevant_feature} provides the additional experiments on Textured MNIST.
Sec.~\ref{sec:training_details} provides the details of the models used in the experiment and step-by-step procedure of the training.

\appendix
\section{Datasets}\label{sec:datasets}
\begin{figure*}[t]
\centering
\begin{tabular}{ccc}
\includegraphics[height=8em]{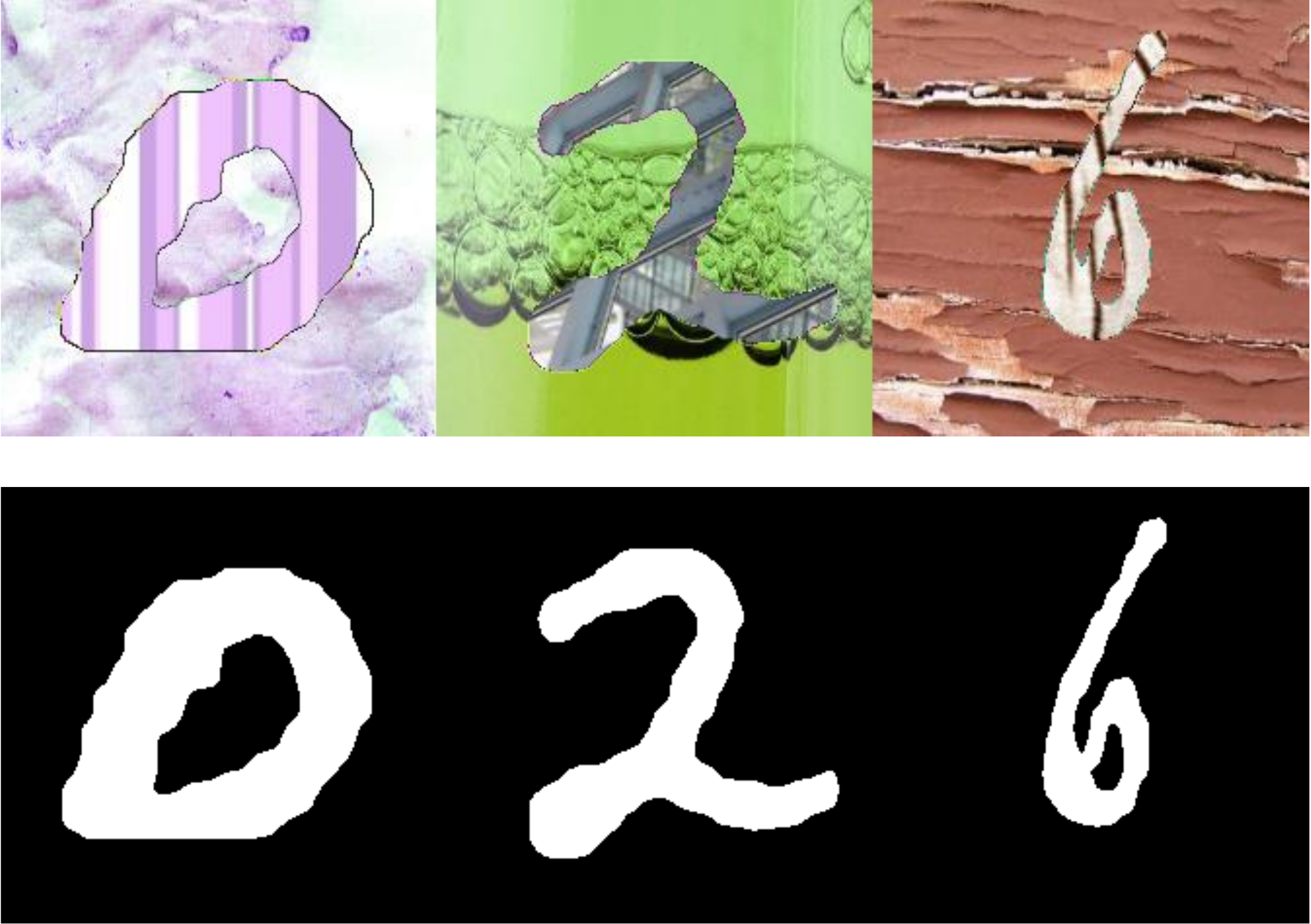} 
&
\includegraphics[height=8em]{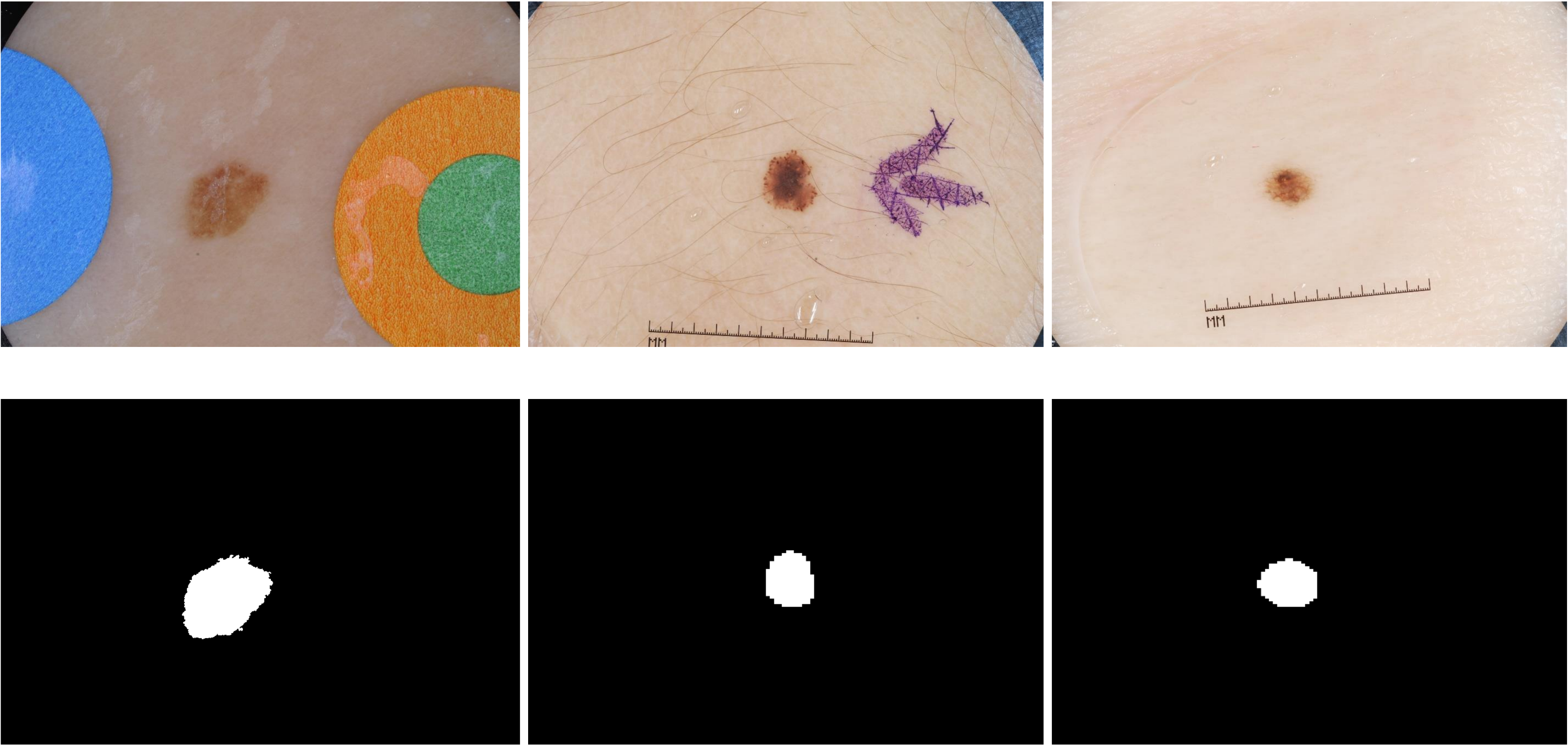}
&
\includegraphics[height=8em]{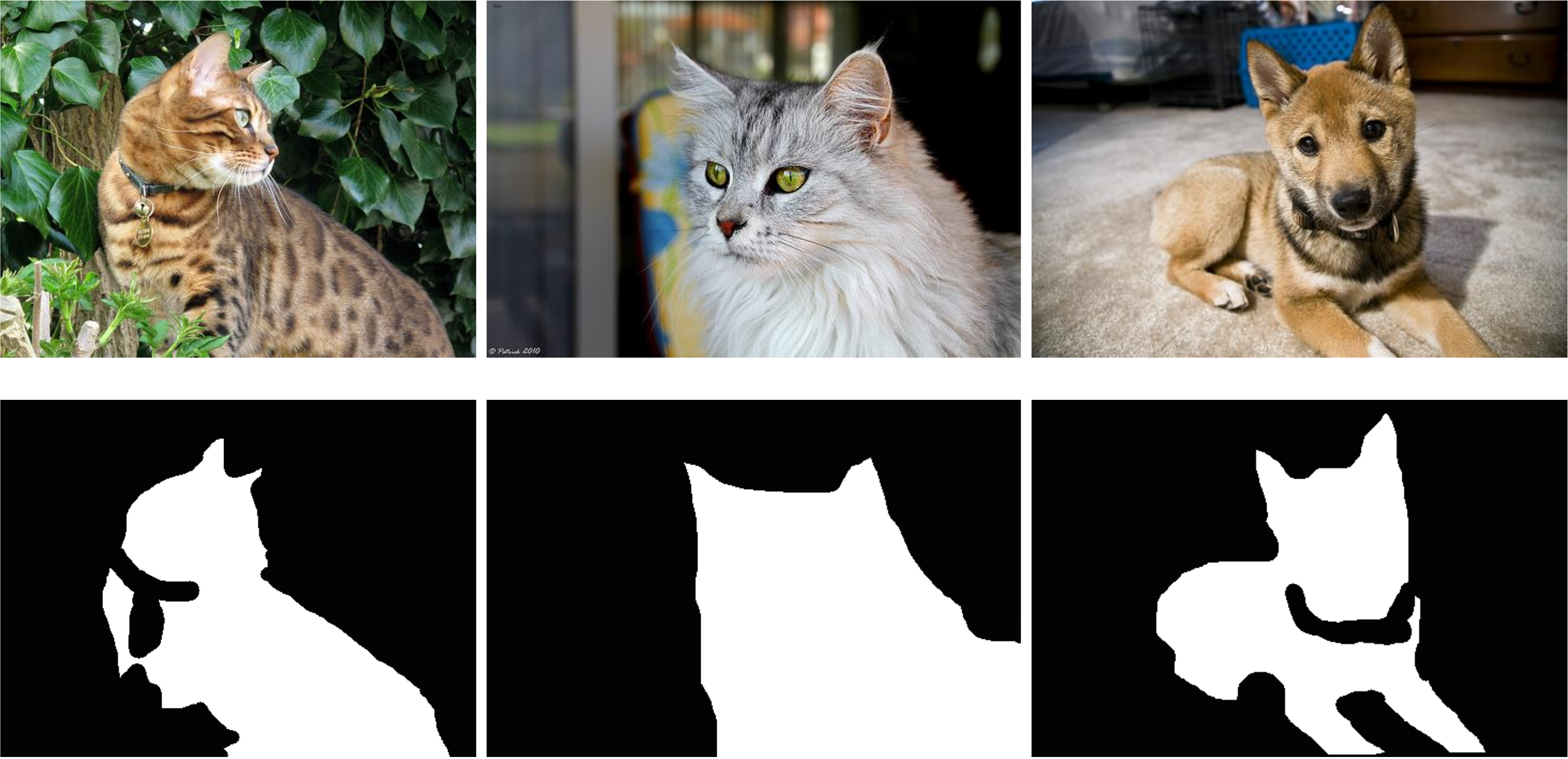}
\\
(a) Textured MNIST & (b) ISIC2017 \cite{codella2018skin} & (c) Oxford-IIIT Pet \cite{oxford_pets}
\end{tabular}
\caption{Examples of the datasets.
The upper and lower rows show the images and the corresponding regional annotations for task-relevant features.}
\label{fig:dataset_examples}
\end{figure*}
\begin{table}[htb]
\caption{Details of the texture classes and the correspondence between the texture classes of DTD and digit classes of MNIST.}
\label{tab:textures}
\includegraphics[width=1\linewidth]{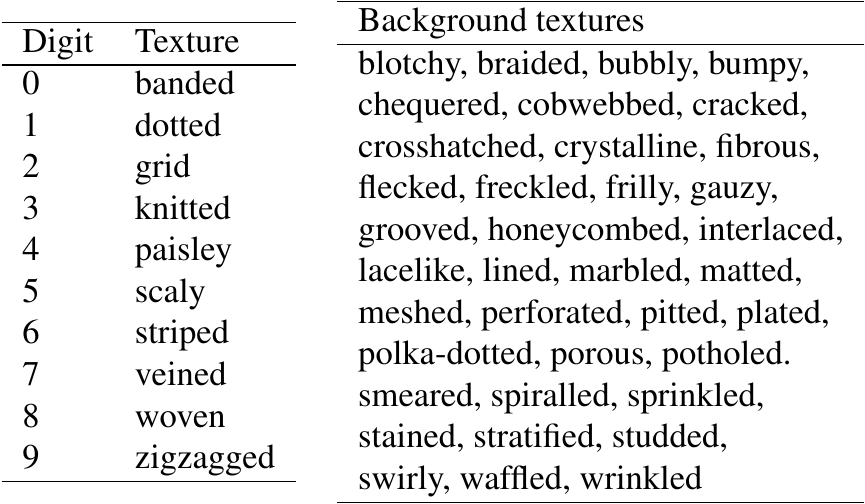}

\end{table}
{\bf Textured MNIST} has 60,000 images for training and 10,000 images for testing, similar to MNIST.
We split the training set into 90\% for training and 10\% for validation.
From DTD, we selected 10 classes for digit texture and the remaining 37 classes for background texture.
The shapes of the digits were used as the regional annotations.
The names of the texture classes and the correspondence between the digit classes and texture classes are shown in~\tabref{tab:textures}
\\
{\bf ISIC2017} contains 2,000 images for training, 150 images for validation, and 600 images for testing.
In addition to the class labels, the images also have segmentations that indicate the regions of the lesions, which we regarded as regional annotations.
\\
{\bf Oxford-IIIT Pet (Pets)} contains 3,680 images for training and 3,669 images for testing.
We randomly split the training set into 90{\%} for training and 10{\%} for validation.
Also, each pixel of each image in this dataset is classified into background, foreground, and indeterminate.
We made the regional annotations by regarding foreground pixels as task-relevant features and the other pixels as not relevant.

\figref{fig:dataset_examples} shows examples of the datasets and corresponding regional annotations explained in Sec.~4.1.

\section{Explanation Regularizer}\label{sec:explanation_regularizer}
\begin{table*}[t]
\centering
\caption{Test accuracy [\%] vs size of training set for pseudo-annotator $n$ on Textured MNIST.
The best accuracy for each $n$ is in {\bf boldface}.}
\label{tab:additional_experiment}
\begin{tabular}{l|llll} \hline
$n$ & ReFF & ReFF (L1) & ReFF (L2) & ABN \cite{embedding_human_knowledge} \\ \hline 
1 & $50.25\pm 2.65$ & ${\bf 50.43\pm 1.19}$ & $45.21\pm 2.45$ & $43.88\pm 2.19$ \\ 
5 & ${\bf 52.09\pm 4.96}$ & $51.47\pm 0.82$ & $44.80\pm 2.53$ & $41.40\pm 2.54$ \\ 
10 & ${\bf 56.73\pm 8.10}$ & $53.39\pm 3.17$ & $45.42\pm 2.10$ & $43.42\pm 2.44$ \\ 
20 & $53.88\pm 1.41$ & ${\bf 55.01\pm 1.55}$ & $47.78\pm 0.28$ & $44.25\pm 3.97$ \\ 
50 & ${\bf 51.81\pm 1.10}$ & $46.27\pm 1.38$ & $47.51\pm 0.57$ & $45.44\pm 4.59$ \\ 
All & $46.44\pm 1.01$ & $43.22\pm 1.08$ & $42.49\pm 0.98$ & ${\bf 53.23\pm 4.77}$ \\ 
\hline 
\end{tabular}
\end{table*}
\newcommand{\camfig}[1]{{\includegraphics[height=5em]{#1}}}
\newcommand{\rs}[1]{\raisebox{2em}{{#1}}}

\begin{figure}
\centering
\begin{tabular}{ll}
\rs{$\mathbf{x}$} & \camfig{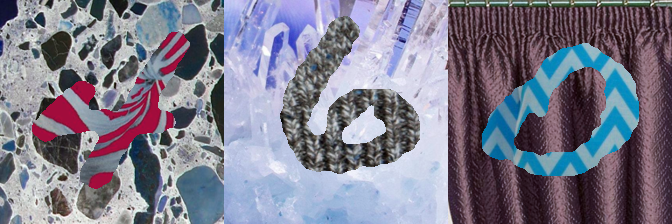} \\
\rs{$\mathbf{s}$} & \camfig{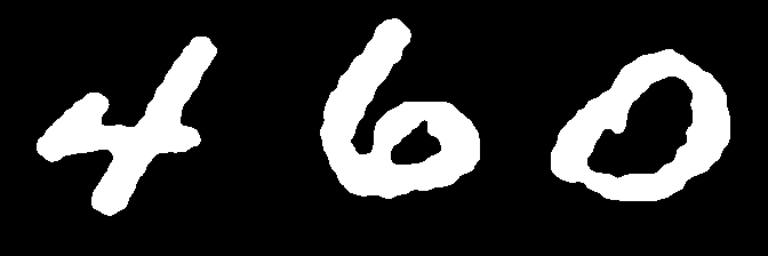} \\
\rs{$n=50$} & \camfig{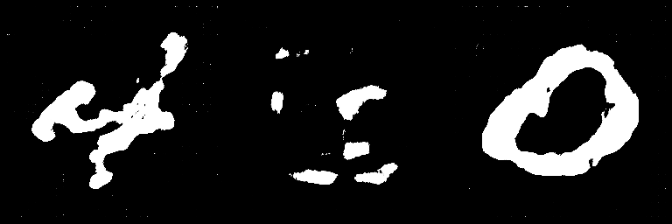} \\
\rs{$n=20$} & \camfig{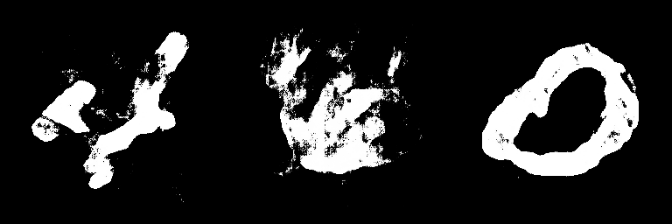} \\
\rs{$n=10$} & \camfig{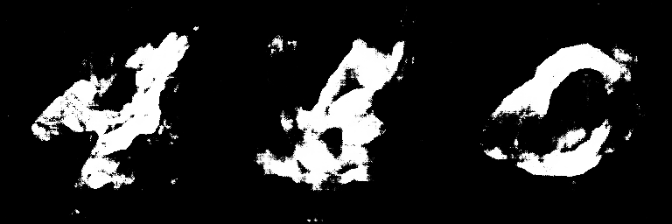} \\
\rs{$n=5$} & \camfig{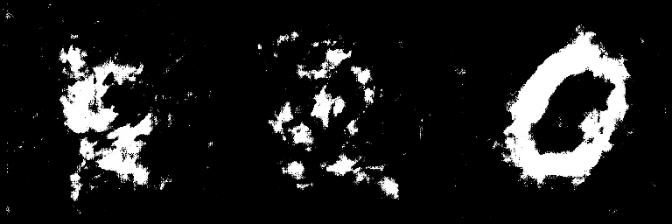} \\
\rs{$n=1$} & \camfig{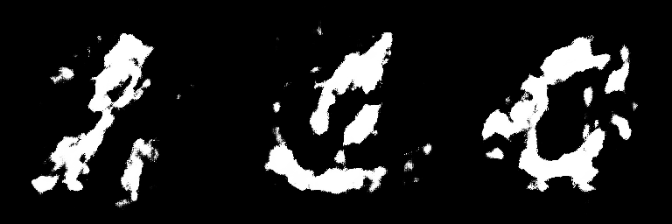} \\
\end{tabular}
\caption{Examples of pseudo-annotations.}
\label{fig:textured_mnist_pseudo-annotation}
\end{figure}

\begin{figure}
\centering
\begin{tabular}{ll}
\rs{$\mathbf{x}$} & \camfig{fig/textured_mnist_cam/textured_mnist_original.png} \\
\rs{$n=\text{`All'}$} & \camfig{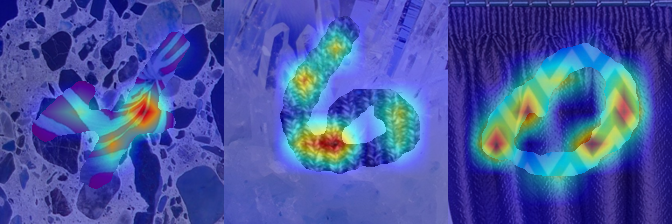} \\
\rs{$n=50$} & \camfig{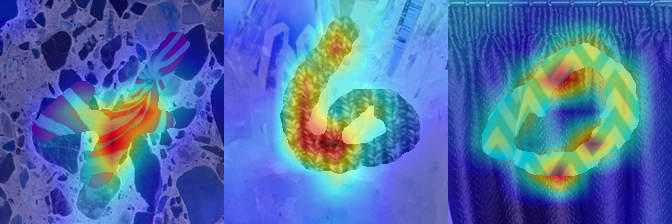} \\
\rs{$n=20$} & \camfig{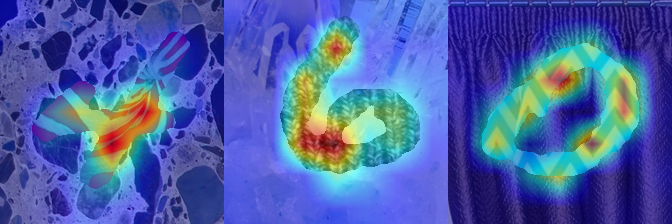} \\
\rs{$n=10$} & \camfig{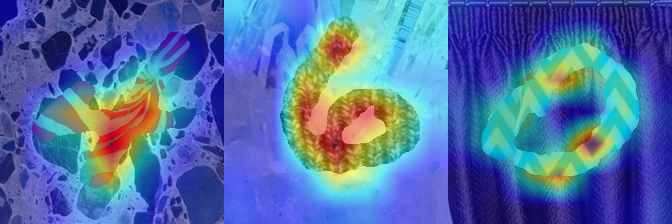} \\
\rs{$n=5$} & \camfig{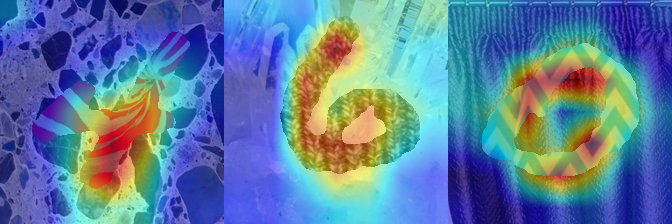} \\
\rs{$n=1$} & \camfig{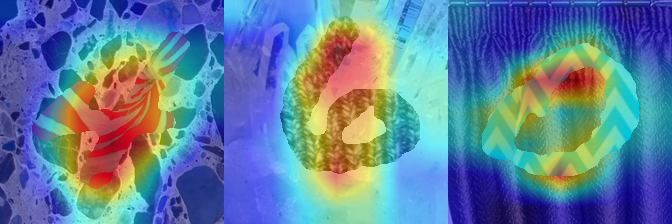} \\
\end{tabular}
\caption{Grad-CAM examples of ReFF.}
\label{fig:textured_mnist_cam_reff}
\end{figure}

\begin{figure}
\centering
\begin{tabular}{ll}
\rs{$\mathbf{x}$} & \camfig{fig/textured_mnist_cam/textured_mnist_original.png} \\
\rs{$n=\text{`All'}$} & \camfig{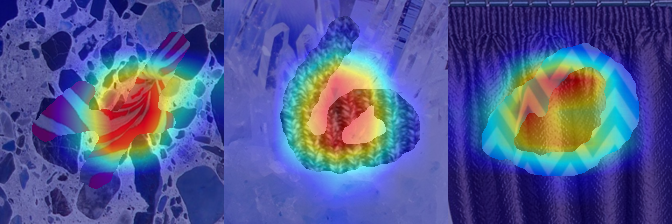} \\
\rs{$n=50$} & \camfig{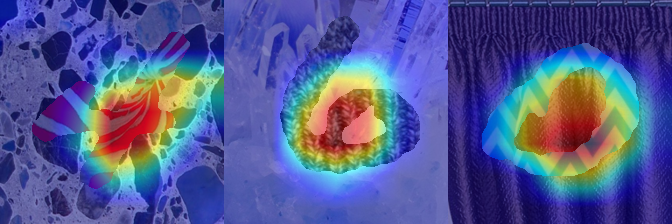} \\
\rs{$n=20$} & \camfig{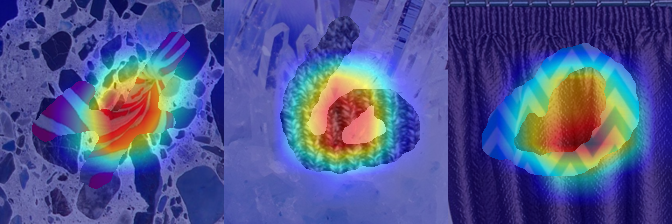} \\
\rs{$n=10$} & \camfig{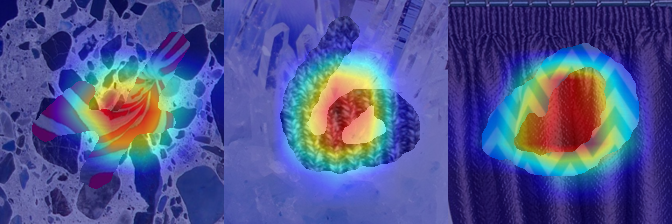} \\
\rs{$n=5$} & \camfig{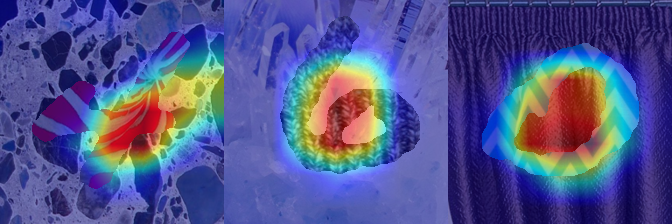} \\
\rs{$n=1$} & \camfig{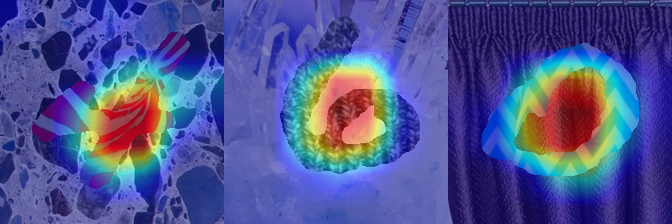} \\
\end{tabular}
\caption{Grad-CAM examples of ReFF (L1).}
\label{fig:textured_mnist_cam_reff_l1}
\end{figure}

\begin{figure}
\centering
\begin{tabular}{ll}
\rs{$\mathbf{x}$} & \camfig{fig/textured_mnist_cam/textured_mnist_original.png} \\
\rs{$n=\text{`All'}$} & \camfig{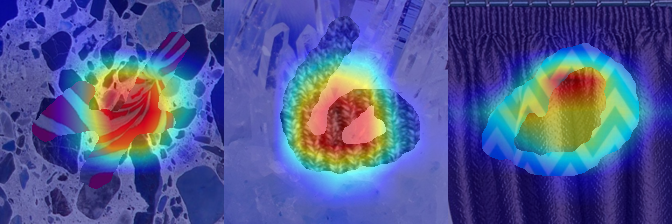} \\
\rs{$n=50$} & \camfig{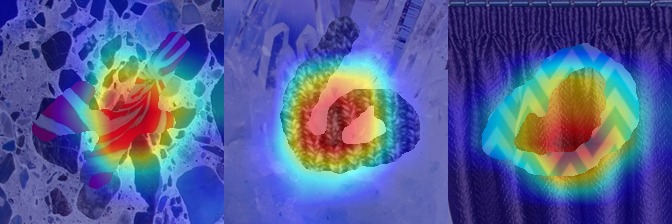} \\
\rs{$n=20$} & \camfig{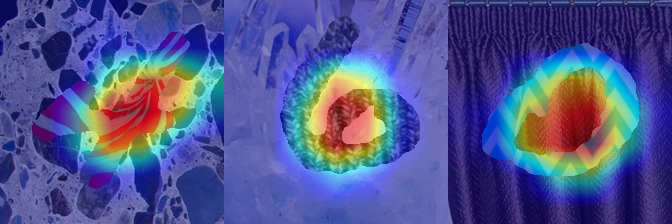} \\
\rs{$n=10$} & \camfig{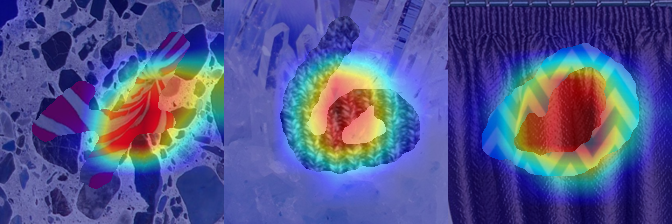} \\
\rs{$n=5$} & \camfig{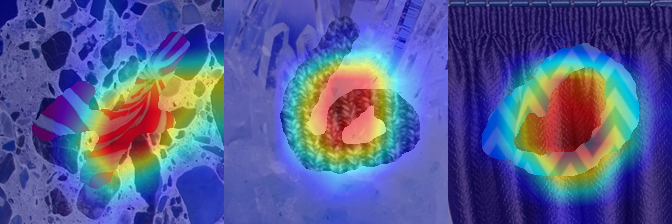} \\
\rs{$n=1$} & \camfig{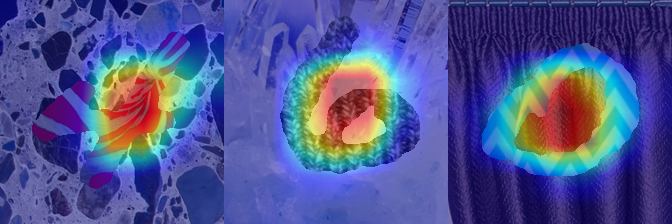} \\
\end{tabular}
\caption{Grad-CAM examples of ReFF (L2).}
\label{fig:textured_mnist_cam_reff_l2}
\end{figure}

\begin{figure}
\centering
\begin{tabular}{ll}
\rs{$\mathbf{x}$} & \camfig{fig/textured_mnist_cam/textured_mnist_original.png} \\
\rs{$n=\text{`All'}$} & \camfig{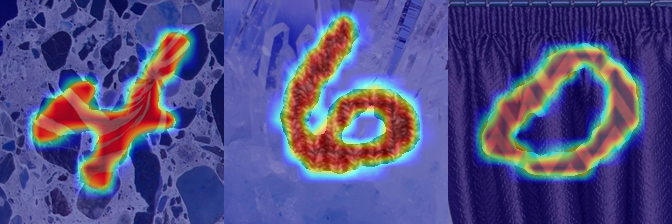} \\
\rs{$n=50$} & \camfig{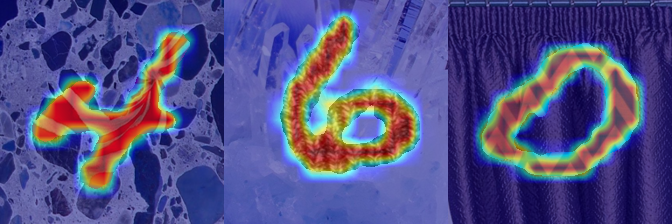} \\
\rs{$n=20$} & \camfig{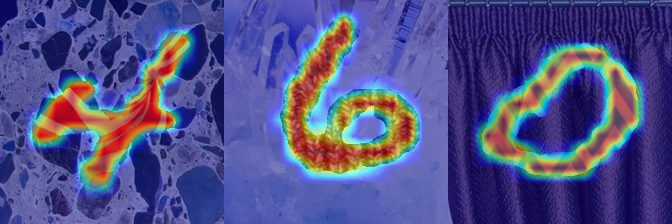} \\
\rs{$n=10$} & \camfig{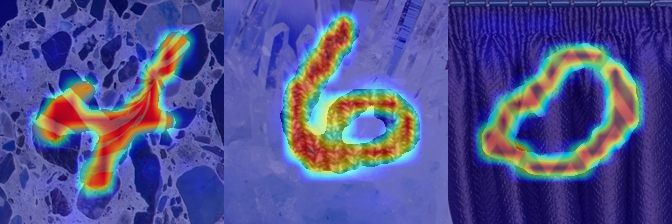} \\
\rs{$n=5$} & \camfig{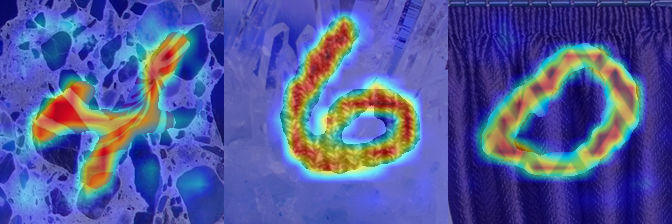} \\
\rs{$n=1$} & \camfig{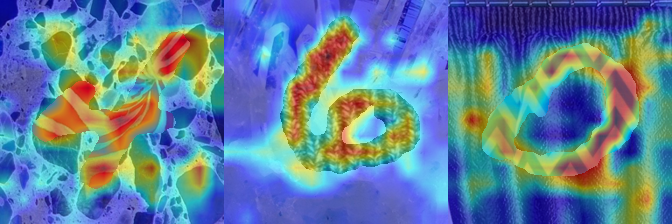} \\
\end{tabular}
\caption{Attention examples of ABN.}
\label{fig:textured_mnist_cam_abn}
\end{figure}

To show the effectiveness of our explanation regularizer, we compared with its variants.
We replaced the explanation regularizer $\mathcal{L}_\text{ReFF}$ described in Sec.~3.1 with the simple L1 or L2 error as follows:
\begin{eqnarray*}
\mathcal{L}_\text{L1} &=& \mathbb{E}_{\mathbf{x},y,\mathbf{s}}\left[ \sum_{l=1}^L w_l \left\| I_l'(\mathbf{x},y,\theta) - \mathbf{s} \right\|_1 \right], \\
\mathcal{L}_\text{L2} &=& \mathbb{E}_{\mathbf{x},y,\mathbf{s}}\left[ \sum_{l=1}^L w_l \left\| I_l'(\mathbf{x},y,\theta) - \mathbf{s} \right\|_2^2 \right].
\end{eqnarray*}
$\mathcal{L}_\text{L1}$ and $\mathcal{L}_\text{L2}$ directly minimize the norm of the differences between a regional annotation and a Grad-CAM, 
whereas $\mathcal{L}_\text{ReFF}$ in Eq.~(3) minimizes the norm of masked Grad-CAM itself.
We set the coefficient $\lambda=10^{-6}$ for $\mathcal{L}_\text{L1}$ and $\mathcal{L}_\text{L2}$, which produced the best accuracy within  $\lambda \in \{ 10^{-8}, 10^{-7}, 10^{-6}, 10^{-5} \}$.
We set $(w_1,w_2,w_3,w_4) = (15,60,250,1000)$, as described in Sec.~4.2.

We also compared ReFF with ABN-based regularization (ABN) \cite{embedding_human_knowledge,Fukui_2019_CVPR}.
While ABN uses only manual regional annotations, we combined it with our pseudo-annotator.
ABN minimizes the L2 error between the attention map $M(\mathbf{x})$ and the regional annotation $\mathbf{s}$:
\[
L_\text{map} =\gamma \| M(\mathbf{x}) - \mathbf{s} \|_2^2.
\]
Note that $L_\text{map}$ regularizes only the attention maps generated from a single attention branch module, 
while $\mathcal{L}_\text{L1}$ and $\mathcal{L}_\text{L2}$ regularize the Grad-CAMs generated from multiple layers.
For training, $L_\text{map}$ was added to the classification loss.
We set the hyperparameter $\gamma=10$, which produced the best accuracy within $\gamma \in \{ 1,10,100 \}$.
We used ResNet-50 as the backbone of ABN.

We trained the models with these methods on TexturedMNIST in the same way as described in Sec.~4.6.
We set the background randomness and digit texture randomness $(p,q)=(0,1)$ for the training and $(p,q)=(1,1)$ for the test.
We also added a pseudo-annotator trained on the subsets of the training dataset.

\tabref{tab:additional_experiment} shows the test accuracy.
When $n\leq 50$ where the pseudo-annotations are sometimes inaccurate, ReFF achieves the highest accuracy in most cases.
This is because ReFF in Eq.~(3) is robust to corruption of the pseudo-annotations 
since it suppresses explanations only outside the regions of task-relevant features as mentioned in Sec~3.1.
In the case of the L1 or L2 error, the models attempt to align their attentions with the corrupted pseudo-annotations strictly and fail to focus on task-relevant features.
On the other hand, when $n=\text{`All'}$ where regional annotations are given for all training data, 
ABN outperformed ReFF because ABN has the specialized module called attention branch that is independent of prediction.
ReFF (L1) and ReFF (L2) almost degrade the accuracy in comparison with ReFF because 
regularizing Grad-CAMs with L1 or L2 error has serious effect on accuracy since the feature maps are shared to output Grad-CAMs and the predictions, unlike ABN.

\figref{fig:textured_mnist_pseudo-annotation} -- \figref{fig:textured_mnist_cam_abn} visualize the pseudo-annotations, Grad-CAMs, and attentions of the trained models.
As $n$ becomes smaller, while the pseudo-annotations become less accurate as shown in~\figref{fig:textured_mnist_pseudo-annotation}, 
the Grad-CAMs of ReFF in~\figref{fig:textured_mnist_cam_reff} highlight the digits.
On the other hand, the attentions of ABN in \figref{fig:textured_mnist_cam_abn} highlight the digits accurately when $n$ is large. But the attentions are corrupted when $n=1$ in contrast to ReFF.
Regarding ReFF (L1) and ReFF (L2), the Grad-CAMs in \figref{fig:textured_mnist_cam_reff_l1} and \figref{fig:textured_mnist_cam_reff_l2} highlight mostly only the central area regardless of $n$ 
because the models fail to cope with the prediction and focusing on task-relevant features at the same time,
since the feature maps are shared to generate the Grad-CAMs and the predictions.

\begin{figure}[tb]
\centering
\begin{tabular}{c}
\includegraphics[width=0.6\linewidth]{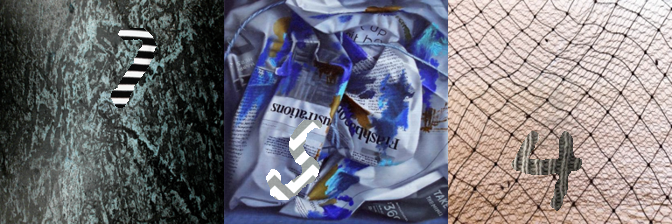} \\
\includegraphics[width=0.6\linewidth]{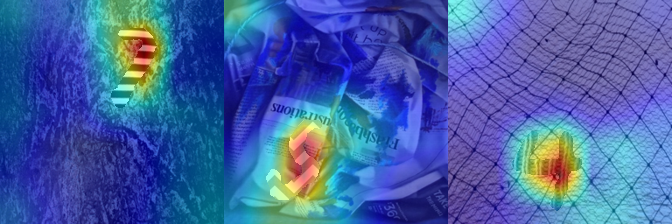}
\end{tabular}
\caption{Grad-CAMs for the model trained with ReFF on Textured MNIST whose digit size is 112.
The upper and lower rows show the input images and corresponding Grad-CAMs.}
\label{afig:cam_reff_112}
\end{figure}

Further, to check whether a model trained with ReFF actually learns the relevant features, not just memorizes the positions of the digits, 
we also visualize Grad-CAMs for the model trained on Textured MNIST whose digit size is 112.
\figref{afig:cam_reff_112} shows the Grad-CAMs.
We can see that the Grad-CAMs highlight the digits regardless of their positions.
Hence, we can say that the model trained with ReFF learns the relevant features.

\section{Effect of Task-relevant Features}\label{sec:experiment_relevant_feature}
\begin{table}[tb]
\centering
\caption{Test accuracy [{\%}] vs $q$ on Textured MNIST ($p=1$ in training dataset).
For each condition, we ran the training three times with different random seeds; 
the standard deviations are shown together.
Note that $p=1$ means that the training dataset does not contain spurious features.}
\label{tab:exp_task-relevant_features}
\begin{tabular}{l|ll} \hline
$q$ & Baseline & ReFF \\ \hline
0.0 & $99.67\pm0.04$ & ${\bf 99.70\pm0.06}$ \\
0.2 & ${\bf 99.60\pm0.06}$ & $99.58\pm0.04$ \\
0.4 & $99.49\pm0.06$ & ${\bf 99.57\pm0.05}$ \\
0.6 & $99.48\pm0.06$ & ${\bf 99.53\pm0.03}$ \\
0.8 & $99.53\pm0.01$ & ${\bf 99.56\pm0.06}$ \\
1.0 & $99.46\pm0.05$ & ${\bf 99.49\pm0.08}$ \\ \hline
\end{tabular}
\end{table}

We evaluated the effect of ReFF under the condition that spurious features does not exist and the task-relevant features vary.
Here, we conducted the experiment on Textured MNIST.
We fixed background randomness $p=0$ and varied the digit texture randomness $q$ in Textured MNIST.
For the training dataset, we fixed the background randomness $p=1$ and varied the digit texture randomness $q$.
For the test dataset, we fixed $p=1$ and set the same $q$ as that of the training dataset.
\tabref{tab:exp_task-relevant_features} shows the test accuracies without spurious features in the training dataset ($p=1$).
The baseline and ReFF achieved almost same accuracy ($>99{\%}$).
This result means that ReFF does not have any negative effect on training when no spurious features exist.

\section{Training Details}\label{sec:training_details}
Here, we detail the settings of the experiments described in Sec.~4.2. \\
{\bf Classification:} The hyperparameter settings are summarized in \tabref{tab:hyperparameters} (a).
We downloaded the initial weights pretrained on ImageNet~\cite{lsvrc2012} via {\tt torchvision} library\footnote{\url{https://pytorch.org/vision/stable/index.html}}.
The optimizer settings, except the weight decay, are the default values in PyTorch \cite{pytorch}. \\
{\bf Pseudo-annotator:} The implementation was done in the same way as pix2pix \cite{pix2pix}.
We set the hyperparameters of the optimizer as shown in \tabref{tab:hyperparameters} (b).
We implemented the discriminator with a 5-layer CNN.
The output channels of the layers numbered 64, 128, 256, 512, and 1.

The training procedure of ReFF is shown in Alg. \ref{alg:reff}

\renewcommand{\algorithmicrequire}{\textbf{Input:}}
\renewcommand{\algorithmicensure}{\textbf{Output:}}

\begin{algorithm*}
\caption{ReFF training with pseudo-annotator}
\label{alg:reff}
\begin{algorithmic}
\Require Regional-annotated training samples $\mathcal{D}_1=\{ (\mathbf{x}_i,y_i,\mathbf{s}_i) \}$,\\ training samples without regional annotations $\mathcal{D}_2=\{ (\mathbf{x}_i, y_i) \}$
\Ensure Trained classifier $F_\theta$
\State{Train pseudo-annotator $G_\phi$ on $\mathcal{D}_1$}
\Repeat
\Comment{Train classifier $F_\theta$}
\State{Sample mini-batch $\mathcal{B}$ from $\mathcal{D}_1 \cup \mathcal{D}_2$}
 \State{$(\mathcal{L}_\text{ReFF}, \mathcal{L}_\text{main}) \gets (0,0)$}
 \For{$(\mathbf{x},y,\mathbf{s})$ in $\mathcal{B}$}
  \If{$\mathbf{s}$ is given}
  \Comment{Compute explanation regularizer term}
   \State{Compute loss $l_\text{ReFF}$ with Eq. (3)}
  \Else
   \State{Compute loss $l_\text{ReFF}$ with Eq. (5)}
  \EndIf
  \State{$\mathcal{L}_\text{ReFF} \gets \mathcal{L}_\text{ReFF} + l_\text{ReFF}$}
  \State{$\mathcal{L}_\text{main} \gets \mathcal{L}_\text{main} +  l_\text{main}$}
 \EndFor
 \State{$\mathcal{L}_\text{ReFF} \gets \mathcal{L}_\text{ReFF} / |\mathcal{B}|$}
 \State{$\mathcal{L}_\text{main} \gets \mathcal{L}_\text{main} / |\mathcal{B}|$}
 \State{Update $\theta$ to minimize $\mathcal{L}=\mathcal{L}_\text{main}+\lambda \mathcal{L}_\text{ReFF}$}
\Until{convergence}
\end{algorithmic}
\end{algorithm*}

\begin{table*}[tb]
\centering
\caption{Hyperparameter settings of the experiments.}
\label{tab:hyperparameters}
\begin{tabular}{cc}
(a) Classification & (b) Pseudo-annotator \\
\begin{tabular}{ll} \hline
Optimizer & Momentum SGD \\
Learning rate & 0.01 \\
Momentum & 0.8 \\
Batch size & 32 \\
Weight decay & $10^{-5}$ \\
\# of epochs  & 50 \\ \hline
\end{tabular}
&
\begin{tabular}{ll} \hline
Optimizer & Adam \\
Learning rate & 0.0002 \\
$\beta_1$ & 0.5 \\
$\beta_2$ & 0.999 \\
Batch size & 4 \\
\# of iterations & 30,000 \\ \hline
\end{tabular}
\end{tabular}
\end{table*}

\bibliographystyle{IEEEbib}
\bibliography{supplement}